\documentclass[journal]{IEEEtran}

\ifCLASSINFOpdf
\usepackage[pdftex]{graphicx}
\else
\usepackage[dvips]{graphicx}
\fi

\usepackage{url}
\usepackage{multirow}
\usepackage{subfig}

\usepackage[T1,OT1]{fontenc}
\DeclareTextCommand{\DJ}{OT1}{ \raisebox{-0.1ex}{\scalebox{0.75}[1.4]{--}}\kern-.4em D}

\usepackage{diagbox}

\usepackage{todonotes}
\usepackage{soul}

\begin{document}

\title{Open-Ended Evolution\\for Minecraft Building Generation 

\thanks{This project has received funding from the European Union’s Horizon 2020
programme under grant agreement No 951911.}
}

\author{\IEEEauthorblockN{Matthew Barthet,
Antonios Liapis,
Georgios N. Yannakakis\\
\IEEEauthorblockA{Institute of Digital Games, University of Malta, Msida, Malta.\\
Email: \{matthew.barthet, antonios.liapis, georgios.yannakakis\}@um.edu.mt}}}

\markboth{IEEE Transactions on Games}
{Deep Learning Novelty Exploration for Minecraft Building Generation}

\maketitle

\begin{abstract}
This paper proposes a procedural content generator which evolves Minecraft buildings according to an open-ended and intrinsic definition of novelty. To realize this goal we evaluate individuals’ novelty in the latent space using a 3D autoencoder, and alternate between phases of exploration and transformation. During exploration the system evolves multiple populations of CPPNs through CPPN-NEAT and constrained novelty search in the latent space (defined by the current autoencoder). We apply a set of repair and constraint functions to ensure candidates adhere to basic structural rules and constraints during evolution. During transformation, we reshape the boundaries of the latent space to identify new interesting areas of the solution space by retraining the autoencoder with novel content. In this study we evaluate five different approaches for training the autoencoder during transformation and its impact on populations’ quality and diversity during evolution. Our results show that by retraining the autoencoder we can achieve better open-ended complexity compared to a static model, which is further improved when retraining using larger datasets of individuals with diverse complexities.
\end{abstract}


\IEEEpeerreviewmaketitle

\section{Introduction}\label{sec:introduction}

Early works on computational creativity approached the task of content generation by searching the problem space using evolutionary algorithms and an extrinsic objective function \cite{Search_Based_PCG}. As the problem spaces and fitness functions become more complex, however, such algorithms tend to suffer from premature convergence to suboptimal solutions. To overcome this issue, modern creative artificial intelligence (AI) algorithms have shifted toward abandoning objectives altogether and using novelty search to reward solutions based on their diversity \cite{Novelty_Search}. Whilst simple novelty search helps avoid the pitfalls of objective functions, it often still suffers from its own search biases and population convergence \cite{Delenox}, and does not guarantee that solutions exhibit desired qualities. Open-endedness in evolution through novelty \cite{Banzhaf_OE} is an important topic in computational creativity and artificial general intelligence, and is at the center of this project.

Recent works on achieving open-endedness in procedural content generation (PCG) have focused on using an intrinsic definition for novelty which is defined on the system's own output. Intrinsic motivation (IM) in artificial systems has been identified as an important factor for achieving higher forms of open-ended creativity, and in situations where the objective of the system is difficult to model \cite{Guckelsberger_IM}. 
The Deep Learning Novelty eXplorer (DeLeNoX) algorithm \cite{Delenox} approaches an IM novelty function by assessing novelty in terms of a higher-level representation, determined by an autoencoder (CNN). 
By allowing the generator to adjust its measure for novelty according to its own observations, DeLeNoX is able to continually adapt its focus to search beyond its current biases, a concept which is critical for achieving open-ended evolution. More recent works have applied this intrinsically motivated approach to a variety of tasks, such as 2D artifacts \cite{hagg2021expressivity, nguyen2016understanding}, interesting behaviors in robots \cite{cully2019autonomous} and search space illumination \cite{gaier2018dataefficient}. Our approach builds upon DeLeNoX, expanding the algorithm on more complex 3D structures.

\begin{figure}[t]
\centerline{\includegraphics[width=0.95\columnwidth]{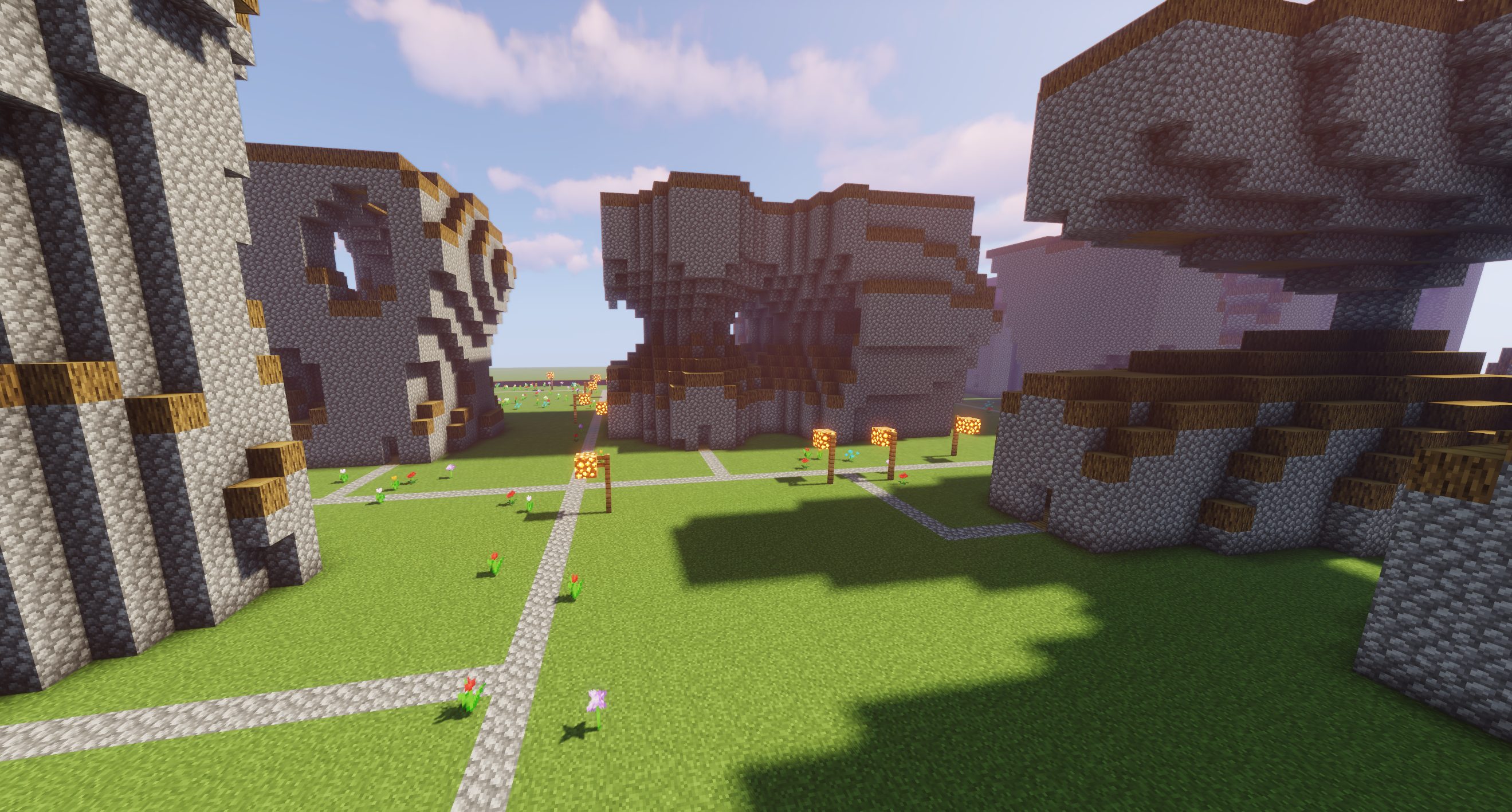}}
\caption{In-game representation of structures generated by our approach, organized into a basic settlement.}
\label{fig:minecraft}
\vspace{-5pt}
\end{figure}

In this paper we propose a computational system designed to autonomously create interesting Minecraft buildings using an intrinsic and open-ended definition of novelty. Sandbox games such as Minecraft \cite{Minecraft} are arguably the perfect canvas to illustrate an artificial system's creativity: their open-ended gameplay allows the player to create any structure that can be expressed as a set of voxels. On a high level, the system alternates between phases of exploration and transformation. During exploration, the latent space defined by the current autoencoder is explored as thoroughly as possible by applying constrained novelty search \cite{Novelty_Search} to neuroevolution \cite{NEAT}. Repair functions ensure individuals abide to a set of basic rules for buildings, in place of an objective function. During transformation, the most novel individuals from each population form a training set to retrain the autoencoder, modifying the latent space (and distance function) and opening up new areas of the solution space to explore. The new autoencoder is used for the next iteration of the algorithm which can continue until stopping criteria are met.

Our findings suggest that redefining the latent space using novel data with a diverse range of structural complexity improves the generator's ability to find more complex and novel features in its output. To our knowledge this approach remains untested for generating complex 3D structures such as Minecraft buildings. Current research into PCG for Minecraft focuses on the goals laid out in the Generative Design in Minecraft Competition \cite{GDMC} to design settlements and buildings, without emphasizing creativity in the generation process. The buildings generated are often static, rule-based systems that do not involve any optimization methods at all. We position this work to fill in this space, and potentially inspire more creativity-focused building and/or settlement generators.

\section{Related Work}\label{sec:relatedwork}

This section provides a brief overview of computational creativity and its link to PCG and games, as well as a look at existing work on assessing novelty in the latent space, and a summary of the state of PCG in Minecraft.

\subsection{Computational Creativity and Games}\label{sec:relatedwork_computationalcreativity}

Computational creativity refers to the study of computer systems which generate content or behave in a manner that an unbiased observer would consider creative, whilst being capable of challenging humans on both a creative and scientific level \cite{Colton_CC}. Whilst early work in this field targeted PCG in isolated, single-faceted domains such as music \cite{Wiggins_Music} and stories \cite{Peinado_Stories}, games have quickly become recognized as the cutting-edge application for the study of computational creativity \cite{CGC}. This is due to their content-intensive and multi-faceted nature which provide rich interactions with the player. Creative systems can also be used as mixed-initiative tools alongside humans \cite{yannakakis2014mixed,yannakakis2018artificial}, helping them to understand and promote the creative process \cite{colton2012computational}.

A critical concept in computational creativity---and artificial life---is \textit{open-endedness} (OE). Open-ended creativity can be informally defined as the ability of a biological or artificial system to keep producing novelty and complexity without exhaustion \cite{Banzhaf_OE}. This definition can be further broken down into three classes (exploratory, expansive, transformative) according to the extent to which the system must redefine its internal models, and its ability to alter the boundaries of the search space to explore new regions and/or dimensions \cite{Taylor_OE}. Modelling and evaluating OE is an active research topic in this field \cite{stepney2021modelling}. Our approach to evaluating novelty in the latent space does not fit ideally with the three classes mentioned above. Whilst our generator could be said to produce expansive novelty in the latent space through transformation, the boundaries and dimensions of the search (phenotype) space remain static, meaning it can only achieve exploratory novelty in the problem space. Achieving higher forms of OE in the phenotype space could involve exploiting the phenotype resolution (expansive) and availability of materials (transformative) to open up new dimensions of the search space.

\textit{Intrinsic motivation} (IM) refers to an individual's desire to explore new behaviors out of an internal curiosity in the outcome \cite{Harlow_IM}, acting freely without a separate motive or external influence \cite{Ryan_IM}. IM in artificial systems has been used in situations where the objective is difficult to define, in situations where transformative open-ended novelty is desired, and as a method for increasing creative autonomy \cite{Guckelsberger_IM}. These properties directly align with our goals for implementing a Minecraft building generator with open-ended complexity and creativity. Our proposed generator periodically redefines its novelty function according to interesting individuals it has generated in the past (without any external influence), satisfying the definition for IM given above. Through our evaluation we discuss the issues and benefits of largely ignoring the quality (extrinsic factors) of candidates and solely focusing on their novelty as a fitness.

Within games, PCG is commonly tackled as a search-based task \cite{Search_Based_PCG}, treating the problem as an optimization task by evolving a population of artifacts according to a desired fitness function which could involve aspects of gameplay behavior or experience \cite{yannakakis2011experience}. PCG for quality-diversity \cite{PCG-QD} is a recent search-based approach which uses quality-diversity algorithms (e.g., MAP-Elites \cite{mouret2015illuminating} with novelty search \cite{Novelty_Search} or surprise search \cite{Surprise_Search}) to preserve diversity in the behavior space whilst improving their performance according to a desired fitness function. On the other hand PCG via Machine Learning \cite{PCG-ML} (PCGML) uses machine learned models which are trained on existing game content to generate new data, removing the need for complicated code to express the designer's intentions for the target domain. The models created through PCGML algorithms can be used to recognize and repair infeasible components of individuals \cite{summerville2016super}, predict properties of their content \cite{jain2016autoencoders}, and compress the content into fewer dimensions using an autoencoder \cite{volz2018evolving}. We cover uses of PCGML and other algorithms specifically for Minecraft in section \ref{sec:relatedwork_minecraft}. Our approach combines the evolutionary principles of PCG-QD, by targeting novelty in the latent space and ensuring quality through constraints, while also leveraging learned representations not for generation directly---as in PCGML---but for the evaluation of novelty in a search-based paradigm.

\subsection{Novelty in the Latent Space}\label{sec:relatedwork_delenox}
Assessing an individual's novelty through its latent representation is a viable method for achieving OE as the boundaries of the latent space can be expanded by retraining the encoder model. 
Furthermore, through this approach novelty is calculated on the high-level patterns encoded in the latent representation, providing a more meaningful measure regarding the (current) possibility space.
The deep-learning novelty explorer paradigm \cite{Delenox} (DeLeNoX) leverages this concept by cycling between phases of exploration and transformation. In exploration, the generator uses the current autoencoder to thoroughly search the current latent space. In the first implementation of DeLeNoX \cite{Delenox}, this was accomplished through neuroevolution of 2D spaceships with constrained novelty search \cite{liapis2015ecj} in the latent space. During transformation, the autoencoder is re-trained using the latest novel individuals from the previous exploration phase. This effectively readjusts the boundaries of the latent space and opens up new potential areas of interesting individuals, improving the generator's ability to conduct an open-ended search. 

The high-level approach of DeLeNoX has since been adapted and extended for a number of other use cases. A similar study to DeLeNoX generates interesting 2D shapes using variational autoencoders \cite{hagg2021expressivity}, and demonstrated its effectiveness in generating diversity over directly evolving individuals' latent vectors. Similar methods have been used for autonomous discovery of novel robot behaviors \cite{cully2019autonomous}, 
data efficient search space illumination \cite{gaier2018dataefficient}, the generation of interesting images \cite{nguyen2016understanding} and Super Mario Levels \cite{shu2021experience} with open-ended novelty or other diversity measures. Another DeLeNoX-based approach has been used for visual DOOM playing using neuroevolution and a latent representation of raw pixel data \cite{alvernaz2017autoencoder}. To the best of our knowledge, the core principles of DeLeNoX---as explored in this paper---have never been applied to generate novel 3D content such as Minecraft buildings.

\subsection{Minecraft Settlement Generation}\label{sec:relatedwork_minecraft}

Minecraft's \cite{Minecraft} open-ended, sandbox design makes it a brilliant canvas for players and computers to express creative behavior by building structures that adapt to the world around them. Given these strengths, it has become a popular testbed for research, most notably in the Generative Design in Minecraft Competition \cite{GDMC} (GDMC). The GDMC encourages people to design AI programs for Minecraft that achieve human-like creativity. The task presented by the competition is to develop an autonomous Minecraft settlement generator that produces functional settlements that are capable of adapting to their environment, whilst remaining aesthetically pleasing and evoking an interesting narrative to the viewer. Due to the complexity of these challenges, the first round of contestants did not focus on creativity of the individual buildings, opting for simpler rule-based constructive generators \cite{GDMC_Year1}. 

Content generators have been developed for Minecraft outside the objectives related to the GDMC. The chronicle challenge \cite{salge2019generative} addresses computational storytelling by requiring participants to generate meaningful stories for their settlements. Cellular automata have been used to create realistic buildings with a basic design for the interior volume  \cite{CA_Minecraft}. Neural cellular automata have been used to create structures such caves, buildings, and trees with increasing complexity and ability to regenerate and repair themselves \cite{NCA_Minecraft}. Generation of the world itself has also been researched through works such as World-GAN \cite{World_GAN} which attempts to address the shortcomings of the static world generator bundled with Minecraft through generative adversarial networks. We position this work to fill the gap in the research space for interesting and creative buildings for Minecraft settlements.

\section{Methodology}\label{sec:methodology}
This paper proposes an autonomous Minecraft building generator with open-ended creativity and complexity by evaluating novelty in the latent space. We adapt the core principles of DeLeNoX \cite{Delenox} for this task, describing our approach to building representation and generation in Section \ref{sec:methodology_generation}, the exploration phase in Section \ref{sec:methodology_exploration}, and transformation of the autoencoder in Section \ref{sec:methodology_transformation}.

\subsection{Representation and Generation}\label{sec:methodology_generation}
\begin{figure}[!tb]
\centerline{\includegraphics[width=\columnwidth]{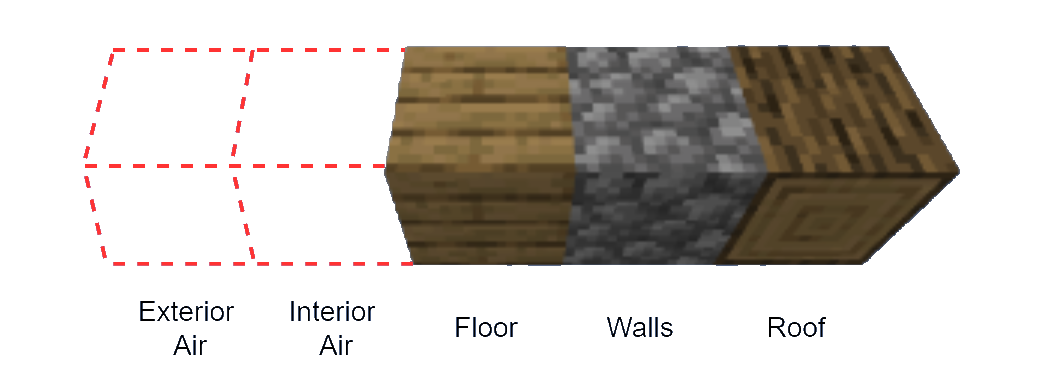}}
\caption{Mapping of our voxel types to in-game materials, useful for reconstructing the generator's output inside Minecraft. Each voxel in the lattice may encode one of five materials: outdoor air, indoor air, floor, wall and roof. Air voxels are empty: the distinction between outdoor and indoor is useful for calculating the building's volume and other constraints.}
\label{fig:voxels}
\end{figure}

In our approach, we represent candidate solutions as three-dimensional arrays of voxels: each voxel encodes a material ID for its location. This aligns with Minecraft's in-game representation of data and allows us to transfer the content generated directly to the game using useful tools such as MCEdit \cite{MCEdit}, an example of which can be seen in Figure \ref{fig:minecraft}. It should be noted that our representation does not directly use Minecraft materials such as wood and stone but rather encodes architectural properties of the space that are needed for the evaluation of the building. We provide the mapping of our voxel types to Minecraft materials for generating in-game models in Figure \ref{fig:voxels}. The material in each voxel is represented as a one-hot encoded vector of binary values. The chosen resolution for the lattices was 20x20x20 voxels split into 5 channels (one for each material), similarly to how an RGB image is represented. The resolution of the lattices and number of materials available to the generator were kept constant and could not be altered during the evolutionary process.

\begin{figure*}[!tb]
\centerline{\includegraphics[width=0.9\textwidth]{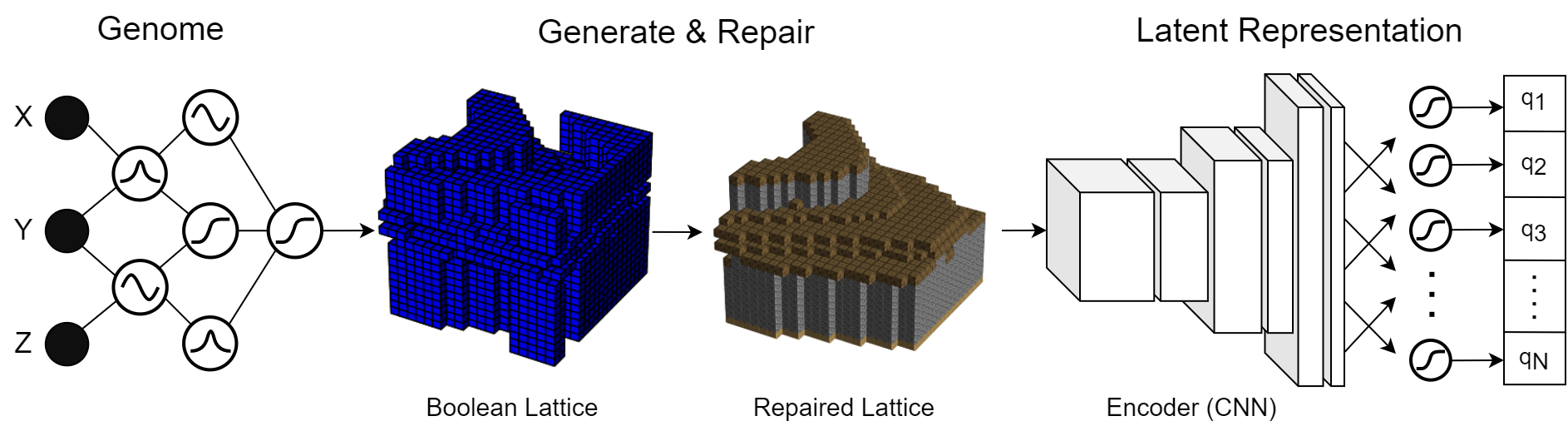}}
\caption{Generating 3D structures using CPPNs and compressing them into latent vectors using an encoder.}
\label{fig:generation}
\end{figure*}

The genetic representation for individuals is a critical design decision when creating search-based content generators, especially when aiming for open-ended evolution, as this choice can limit the maximum complexity achievable. We use compositional pattern producing networks (CPPNs) \cite{stanley2007compositional} to encode the 3D lattice, due to their ability to generate aesthetically pleasing structures. CPPNs are also capable of encoding a phenotype of any resolution, providing an ideal balance between representational power and strong computational efficiency. CPPNs have been proven useful for constrained evolution of 3D voxel-based structures \cite{gravina2018fusing} for robot control. The CPPN iterates over all 3D coordinates of the 3D lattice (using $x, y, z$ as input) and its output is a boolean value that determines whether the voxel should be filled in that location in the lattice. Figure \ref{fig:generation} depicts the process of generating lattices from genomes, ensuring they follow certain rules through repair and constraints, and compressing them into a latent vector using an encoder. The autoencoder is fed the material lattice, rather than the boolean lattice, to ensure material data is encoded in the latent vector and considered when assessing the novelty of a structure.

As noted above, the output of the CPPN is a boolean lattice of empty or filled voxels (the architectural ``hull'' of the building). A set of repair functions remove problematic sections of this boolean hull and convert it into the material lattice (with 5 materials described above). 
Repair functions are applied in sequence: 
first, any floating voxels (i.e., not indirectly attached to a floor voxel) are removed using a flood-fill algorithm. Flood-fill works by picking a starting point from the list of floor voxels and recursively visiting its neighbors in all directions, marking solid voxels and ignoring empty voxels. This process is repeated until every floor voxel has been flooded from, discarding the solid voxels not visited during the process.
Next, only the largest detected structure is kept, removing any voxels detached from this main structure.

After the repair process, materials are assigned to each filled voxel based on a rule-based system in the following order: a voxel at the absolute bottom of the coordinate system becomes floor, a voxel with an empty voxel above it becomes roof, voxels surrounded on all sides by filled voxels become interior air, finally any remaining empty voxels are marked as exterior air and remaining filled voxels become walls. This process also carves out the interior of the structure, marking the voxels as interior air which is a helpful distinction for tracking the interior volume of the building. A final constraint is checked at this point: a structure must contain an valid location for an entrance. This is checked by iterating over the floor voxels and checking if there is a wall at least three voxels high for an entrance (in any of four possible rotations) adjacent to interior air voxels to ensure the area is traversable.

\subsection{Exploration}\label{sec:methodology_exploration}
The role of the exploration phase is to thoroughly search the current boundaries of the latent space for interesting candidate solutions. Our approach to this phase follows DeLeNoX, where neuroevolution of augmenting topologies \cite{stanley2002neat} (NEAT) is used to evolve several populations of CPPNs using constrained novelty search in the latent space. The use of NEAT is ideal for open-ended evolution as it is capable of producing unbounded genetic complexity whilst also preserving genetic diversity through speciation. During exploration, the generator runs NEAT on a set of separate populations and the current autoencoder.  

At the start of each generation of NEAT, the process described in Section \ref{sec:methodology_generation} is executed to create a population of lattices from the genome pool. Due to our single constraint, any individuals that fail it are discarded and replaced during the next round of reproduction. The feasible lattices are compressed using the current autoencoder (as described in Section \ref{sec:methodology_transformation}) resulting in a second `snapshot' of the population containing their latent vectors. These latent vectors are used to calculate the novelty scores for each individual which is done using the average Euclidean distance to the $k$-nearest neighbors in the latent space (Eq. \ref{eq:novelty}).
\begin{equation}
\label{eq:novelty}
    n(i) = \frac{1}{k} \sum^{k}_{m=1} \sqrt{ \sum^{D}_{n=1} (q_n(i) - q_{n}(\mu_m))^2}
\end{equation}
\noindent where $n(i)$ is the novelty score of individual $i$, $k$ is the number of neighbors in the current population and the novel archive that are considered for novelty ($k=15$ in this paper), $D$ is the number of dimensions in the latent vector and $q_n(i)$ is the value of the latent vector at position $n$ when individual $i$ is provided as input to the autoencoder. 

A novelty archive is maintained for each population to ensure current individuals are compared to previous novel examples. At the end of each generation of NEAT, the $\alpha$ most novel individuals are inserted into the novelty archive provided they are unique (no identical matches in the archive). The novelty archive contains both the lattice and latent representation of novel individuals, as these will be needed during transformation. At the end of each generation, we record a number of statistics on the properties of buildings generated (bounding box size, symmetry, instability and surface area) for our evaluation. These properties could be used to implement more complex feasibility constraint functions in the future. 

\begin{figure*}[!tb]
\centerline{\includegraphics[width=0.9\textwidth]{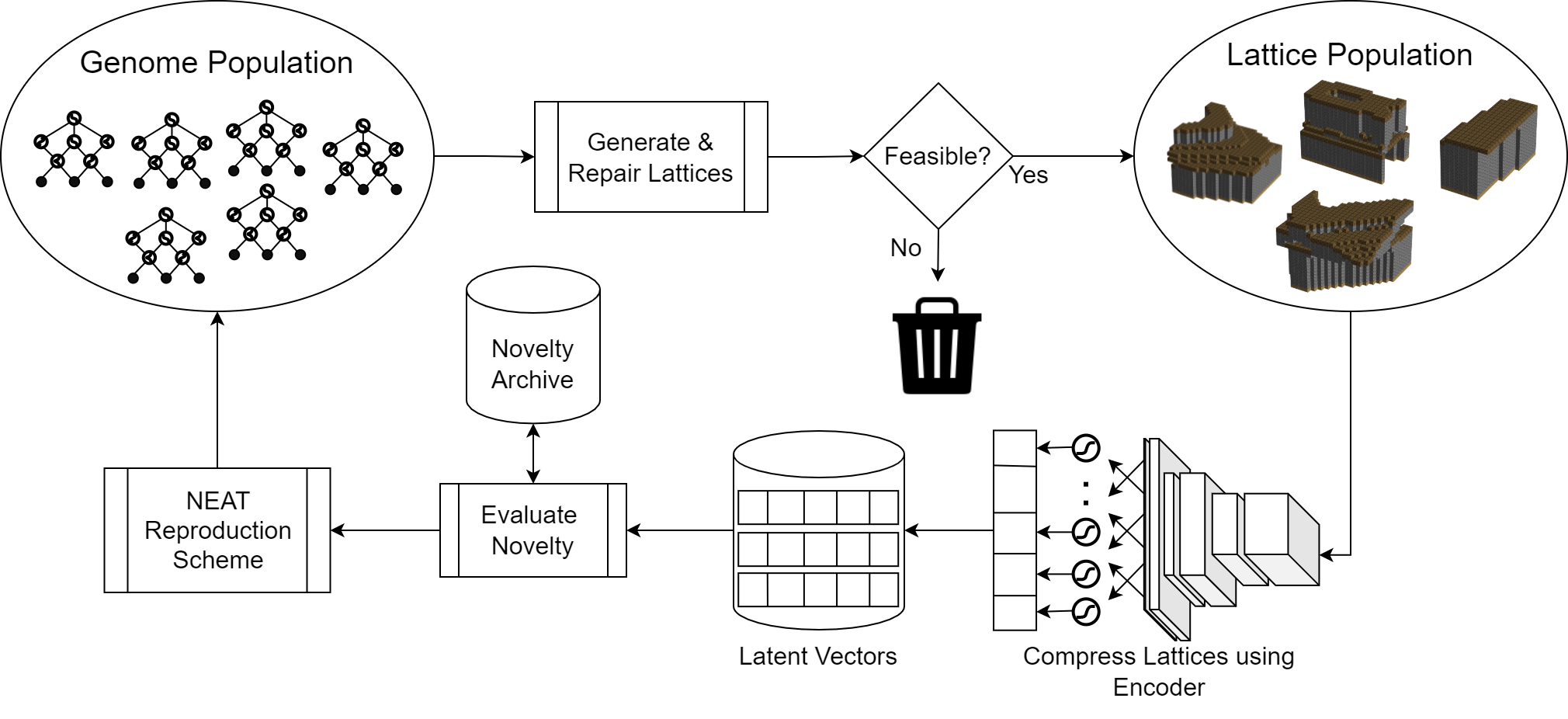}}
\caption{Overview of the exploration phase.}
\label{fig:exploraiton}
\end{figure*}

\subsection{Transformation}\label{sec:methodology_transformation}
As shown in Eq.~(\ref{eq:novelty}), exploration takes place in the latent space by calculating the novelty scores based on each individual's latent vector. Whilst this exploration is capable of effectively searching the current latent space for novel individuals, it is restricted by the boundaries of the manifold. These boundaries are caused by the model that compresses the lattices into latent vectors which in our case takes the form of a 3D convolutional autoencoder. Through exploration alone, the generator is only capable of achieving exploratory novelty \cite{Taylor_OE}, limiting its ability to discover new features in the structures of buildings. By assessing individuals' novelty in the autoencoder's latent space rather than in the lattice space, the generator can shift the boundaries of its current search space by retraining the autoencoder model using individuals it has previously identified as interesting (novel). This results in an intrinsically motivated reward function which is more capable of achieving higher forms of open-endedness (expansive, transformative) compared to a static definition of novelty.

When compressing a 20x20x20 lattice into a 1D vector of 256 values, the encoder model consists of three alternating layers of convolution (of size 3x3x3) and downsampling (of size 2x2x2) which can be seen in Figure \ref{fig:generation}. The weights of this model are assigned during training and are directly responsible for how individuals assess novelty. During transformation, we retrain the encoder model using a training set of individuals taken from the previous exploration phase(s). This retraining process is critical for achieving higher forms of open-endedness, as the new weights of the encoder shift the boundaries of the search space and open up new areas for potentially interesting individuals. From a technical perspective, the retraining allows the encoder to better identify high-level patterns in the building's structure, improving its ability to accurately compress more complex structures.

Transformation begins by retrieving novel individuals from each population of the previous exploration phase(s) and can be done in one of two ways: using  the most novel individuals from each population of the previous exploration phase(s), or using the novelty archives of each population. Both approaches ensure that the autoencoder is transformed using interesting content only which helps to better guide the next exploration phase towards interesting areas of the search space. The two approaches differ in that the former assembles training sets from the final populations of the exploration phase(s) while the latter contains individuals deemed novel throughout evolution. The autoencoders are retrained from scratch using the assembled training set using the categorical cross entropy loss function \cite{zhang2018generalized}. The novelty archives collected so far are updated using the new encoder, assigning new latent vectors for past novel individuals as this will impact how novelty finds neighbors in the archive via Eq.~\ref{eq:novelty}.

\section{Experimental Protocol}\label{sec:experiment}
Assessing the creativity and open-endedness of content generators and the quality of their products is challenging due to the subjectivity of such notions. Since novelty scores are not directly comparable across experiments, we assess the diversity of the structures generated by comparing them at a voxel level.  We use Kullback-Leibler (KL) divergence as our metric for assessing voxel diversity which has proven efficient for comparing game levels \cite{lucas2019tile,shu2021experience}. We also measure the correlation between this KL divergence measure and each experiments' distance measure in the latent space, as this provides an insight into the regularization of the latent space and the novelty function's ability to group meaningfully similar individuals together. We also evaluate the reconstruction accuracy of the autoencoders which directly quantifies the model's ability to identify high-level patterns in the buildings and therefore discover more meaningful novel features. Finally, we provide a qualitative comparison between experiments by visualizing the structures generated and observing the differences in complexity and patterns found to be novel.  

\begin{table*}[tb]
\caption{Reconstruction error (\%) of the final populations of each experiment, based on the final autoencoder (at the end of the 10th round of exploration). Results are averaged from 10 populations tested on the same autoencoder. The last column shows the average reconstruction error for all final populations of all five experiments when using the autoencoder produced by the experiment of that row.}
\centering
\begin{tabular}{|l||c|c|c|c|c||c|}
\hline
\backslashbox{AE}{Pop.}
 & Static & Random & Latest Set & Full History & Nov. Archive & Overall\\ \hline
Static & 10.27 $\pm$ 0.82 & 16.03 $\pm$ 2.77 & 14.00 $\pm$ 0.49 & 15.73 $\pm$ 0.64 & 15.5 $\pm$ 1.52 & 14.31 $\pm$ 2.0\\ \hline
Random & 92.61 $\pm$ 0.11 & 92.05 $\pm$ 0.62 & 94.59 $\pm$ 0.25 & 94.65 $\pm$ 0.3 & 92.45 $\pm$ 0.3 & 93.27 $\pm$ 1.0\\ \hline
Latest Set & 21.57 $\pm$ 1.15 & 26.42 $\pm$ 2.92 & 5.93 $\pm$ 0.24 & 12.68 $\pm$ 1.07 & 20.26 $\pm$ 1.15 & 17.37 $\pm$ 6.0\\ \hline
Full History & 7.19 $\pm$ 0.65 & 13.08 $\pm$ 3.00 & 5.07 $\pm$ 0.27 & 4.01 $\pm$ 0.34 & 12.06 $\pm$ 1.34 & 8.28 $\pm$ 3.0\\ \hline
Nov. Archive & 7.91 $\pm$ 0.49 & 11.92 $\pm$ 2.18 & 9.97 $\pm$ 0.39 & 11.56 $\pm$ 0.67 & 10.41 $\pm$ 0.90 & 10.35 $\pm$ 1.0\\ \hline
\end{tabular}
\label{tab:reconstruction_final_pop}
\end{table*}

Since our method focuses on the transformation of the search space through a latent vector when assessing novelty, the experiment explores different ways of training the autoencoder (AE) and includes two baselines. The first baseline is a \emph{static} AE which was trained on the seed populations (which are common across all experiments) and is not retrained during the transformation phase. While all methods in this paper use the static autoencoder for their first exploration phase, every method except \emph{static} uses a new autoencoder after the first transformation phase. The static AE baseline tests whether retraining the autoencoder has any impact on its own compressibility or the quality of generated results. The second baseline is the \emph{random} AE, wherein each transformation phase a new autoencoder is generated with random weights and is used as-is for the following exploration phase. This baseline tests whether guiding novelty search based on the patterns of the past buildings created has any impact (compared to an untrained autoencoder). The remaining three methods use different training sets during the transformation phase. The \emph{novelty archive} AE (NA-AE) combines all novelty archives from each population in the previous exploration phase to form the training set for the autoencoder. The \emph{latest set} AE (LS-AE) combines only the 100 most novel final individuals of the populations in the previous exploration phase into a training set of 1000 individuals, while the \emph{full history} AE (FH-AE) combines the final individuals in every population of every exploration phase so far to train the autoencoder (1000 individuals multiplied by the number of phases so far).
The three tested methods are designed to assess the importance of all previous exploration steps versus only the last one, and whether maintaining a history of the interim novel results during evolution would yield more robust models.

The experiments were run for 10 iterations of the algorithm (exploration phase followed by transformation phase), evolving 10 separate populations of 200 individuals each. Each exploration phase runs 100 generations of CPPN-NEAT, and transformation retrains the autoencoder for 100 epochs, with a batch size of 64, the ``Adam" optimizer and categorical cross-entropy loss function. Novelty was calculated using the average Euclidean distance to the 15 nearest neighbors in the latent space, and up to 3 individuals are inserted into the novelty archive per generation. For these experiments, autoencoders were trained to compress the $20{\times}20{\times}20{\times}5$ lattices into latent vectors of 256 real values. The first iteration of each experiment uses the same set of seed populations (randomly initialized CPPNs with no hidden nodes) which are used to pre-train an autoencoder (the static autoencoder used as baseline). The first exploration phase uses this trained autoencoder and the seed populations to start the process described in this section. This ensures that whilst there is an element of stochasticity to the exploration and transformation processes, all the experiments start from a common population set and encoder model.  

\section{Results}\label{sec:results}

This section attempts to summarize the findings of the experiments described in Section \ref{sec:experiment}, focusing on the autoencoders' reconstruction accuracy experiments (Section \ref{sec:reco}), the diversity of the final evolved buildings (Section \ref{sec:diversity}), and finally providing a brief qualitative comparison between experiments (Section \ref{sec:qualitative}).

\subsection{Reconstruction Error} \label{sec:reco}

\begin{figure}[tb]
\centerline{\includegraphics[width=\columnwidth]{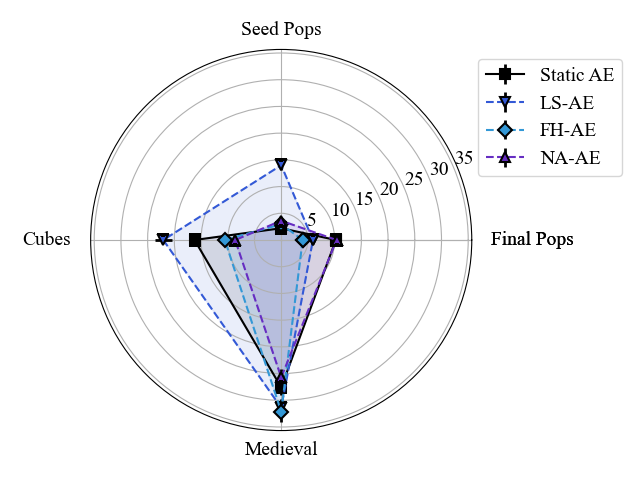}}
\caption{Average reconstruction error (\%) and 95\% confidence interval tested on four datasets of buildings using the final autoencoder from each experiment (after the 10th round of exploration). Results for the Random AE experiment are omitted due to its poor performance (reconstruction error $>$ 90\%) across all tests.}
\label{fig:Reco_Radar}
\end{figure}

Table \ref{tab:reconstruction_final_pop} shows the reconstruction error of the final autoencoders of each experiment, when they reconstruct the final populations of each experiment (and, in the Overall column, all final populations produced by all five methods). On the one hand, we are interested in the ability of the autoencoder to accurately reconstruct individuals evolved based on this autoencoder's notion of novelty (i.e. when AE and pop. are the same). On the other hand, we want to know whether some autoencoders are better at reconstructing buildings in general (i.e. individuals produced by other experiments) as this would indicate that either the autoencoder is more powerful and generalizable or that the buildings produced by different autoencoders and experiments share similar patterns. 

Results in Table \ref{tab:reconstruction_final_pop} indicate that the full history autoencoder (FH-AE) is not only able to reconstruct its own populations more accurately, but is also able to reconstruct all other sets of generated buildings from all experiments (see Overall column). Moreover, it is interesting to note that the latest set autoencoder (LS-AE) is able to reconstruct LS populations, but struggles when facing populations generated by all other methods. Interestingly, the reverse is true with the novelty archive autoencoder (NA-AE) which has comparable or lower reconstruction errors in other populations than the ones it was evolving. While the random (untrained) autoencoder understandably performs very poorly, the static autoencoder is surprisingly robust even when reconstructing the latest populations regardless of experiment. This may indicate that the patterns of the CPPNs that produce the buildings remain concise, regardless of the CPPNs' augmented topologies at later stages. On the other hand, buildings evolved using the static autoencoder are also fairly predictable, as their reconstruction error is low for both the FH-AE and NA-AE.

\begin{figure}[tb]
\centerline{\includegraphics[width=0.9\columnwidth]{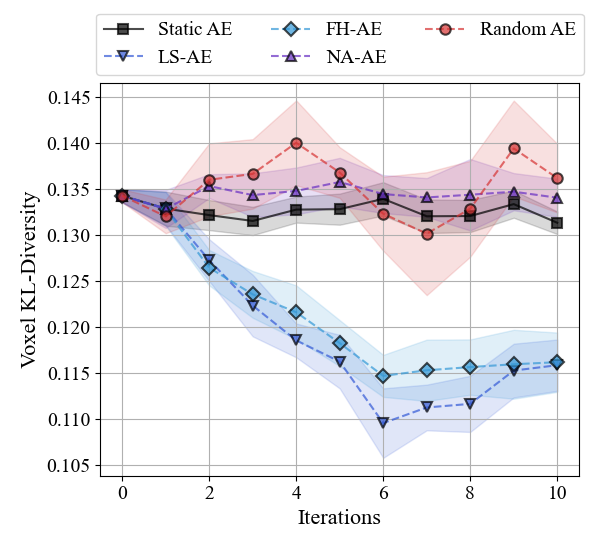}}
\caption{Voxel-based KL Divergence of the populations of each experiment after every round of exploration. Results are averaged across all 10 populations using a 95\% confidence interval. Iteration zero depicts the average diversity of the seed populations.}
\label{fig:KL_Divegence}
\end{figure}
\begin{figure}[tb]
\centerline{\includegraphics[width=0.9\columnwidth]{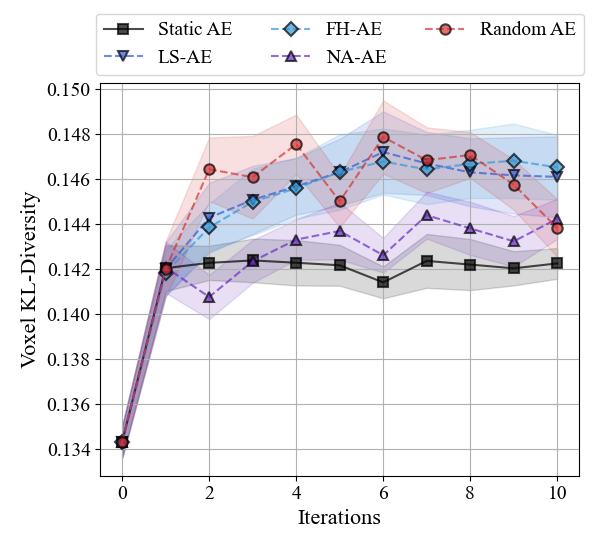}}
\caption{Voxel-based KL Divergence of each experiment's populations from the seed populations used to start evolution. Results are averaged across all 10 populations using a 95\% confidence interval. Iteration zero depicts the average diversity of the seed populations.}
\label{fig:KL_Seed}
\end{figure}

Figure \ref{fig:Reco_Radar} shows the same reconstruction error measure tested across four different datasets to visualize the autoencoders' accuracy across a variety of inputs. The seed and final populations refer to populations at the start (before evolution) and end (end of 10th exploration phase) of each respective experiment. The ``Cubes'' dataset consists of 200 buildings made by randomly generating cuboid hulls of different sizes and applying the repair pipeline to produce material lattices. The ``Medieval'' dataset consists of a population of buildings generated using the ``AHouseV5'' filter by Adrian Brightmoore \cite{AHouseV5} in MCEdit, re-assigning the materials for each voxel through the repair pipeline. Unsurprisingly, the random AE performs very poorly across all four datasets, followed by LS-AE which struggled to reconstruct anything except its own final population. The NA-AE proved to be the most robust model when given completely unseen data, displaying the best reconstruction accuracy for the Medieval and Cubes datasets. As we have seen in Table \ref{tab:reconstruction_final_pop}, the FH-AE shows the best performance on its final populations, though (like the static AE) struggled slightly compared to NA-AE on completely unseen data. The NA-AE seems to benefit from having the largest amount of (and most diverse) training data for transformation compared to the rest of the experiments. 

\begin{figure*}[tb]
\centerline{\includegraphics[width=0.8\textwidth]{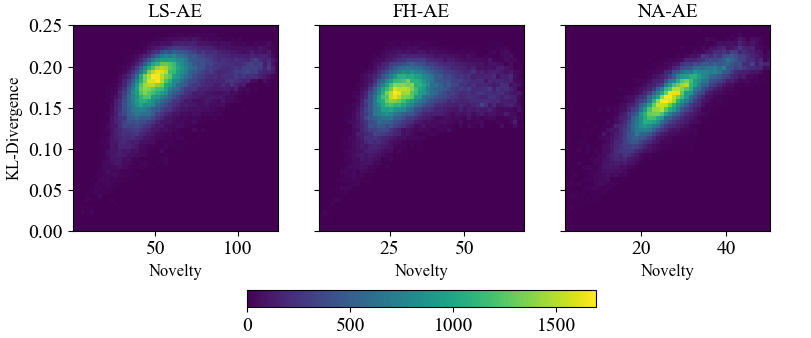}}
\caption{Histogram comparing the distances between individuals in the latent space (i.e: the experiment's novelty measure) and phenotype space (Kullback-Leibler divergence on voxels) for the LS-AE, FH-AE, and NA-AE experiments' final populations. }
\label{fig:NA_Correlation}
\end{figure*}

\subsection{Voxel KL-Divergence} \label{sec:diversity}
Figure \ref{fig:KL_Divegence} shows the progression of the average KL divergence for the populations of each experiment. This was calculated in the voxel space and measures the average KL divergence of each individual to the rest of the population. The results show that FH-AE and LS-AE produce less diverse content in the voxel space over time compared to the rest of the experiments. On the other hand, the static AE produces more diverse content in the voxel space, without varying over time; this is expected as the autoencoder is not retrained between exploration phases. Interestingly, the NA-AE produces a similar trend to the static AE even though it is trained on the largest dataset of individuals during transformation. The random AE produces marginally more diverse content, albeit with a larger deviation which is likely caused by the randomized weights of the autoencoder. It is worth noting that while the LS-AE and FH-AE populations trend downward in voxel-based diversity, it does not in itself mean that the experiments are not generating novel content in the latent space.

Figure \ref{fig:KL_Seed} depicts the evolution of the average KL divergence from the initial seeds (before evolution starts). Similar to the previous metric, this was calculated by computing the average KL diversity of each individual to every individual in the corresponding seed population. The results show that whilst the LS-AE and FH-AE produced the least diverse individuals compared to each other, these same individuals were the most diverse from the initial seeds. The static AE produced the least diverse content for this metric with the least variation over time, indicating that without retraining the autoencoder NEAT struggles to produce significantly different content over time (poor open-endedness). The NA-AE tends to marginally increase in diversity from the seed populations over time, though not to the extent of LS-AE and FH-AE. The random AE produced the most inconsistent trend which is to be expected due to the lack of training for the autoencoder during the transformation phase.

\begin{figure*}[tb]
\centerline{\includegraphics[width=0.85\textwidth]{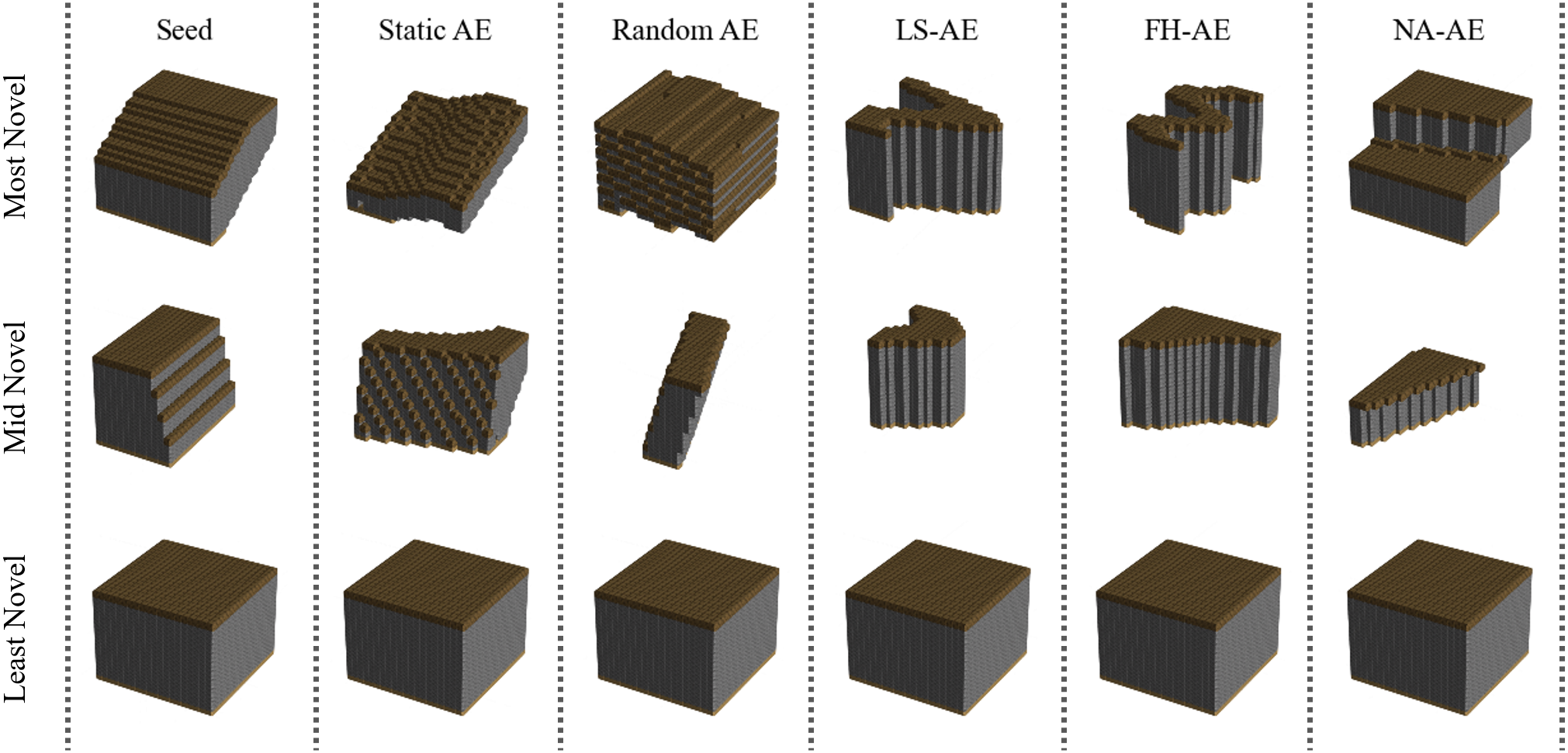}}
\caption{Visualization of individuals from each experiments' final population sorted according to their novelty score (minimum, median, maximum). To evaluate novelty the final autoencoder from each experiment was used (excluding the seed and static AE populations which used the seed model).}
\label{fig:Examples}
\end{figure*}

While autoencoders were trained solely to minimize reconstruction error (i.e., without an explicit regularization component), we visualize the correlation between the novelty score in the latent space and voxel KL divergence in Figure \ref{fig:NA_Correlation}. A regularized latent space is desirable as it ensures similar phenotypes (buildings) are grouped together in the latent space whilst dissimilar individuals are distanced appropriately.  This ensures the validity of the autoencoder's distance measure, as identifying novel individuals in the latent space will result in more meaningfully different individuals in the phenotype space. There is a clear linear correlation between the two diversity measures for the NA-AE experiment's final populations, with a Pearson correlation of 0.84.  However, the LS-AE and FH-AE distance functions both produce significantly weaker correlations between the two measures, with a Pearson correlation of 0.53 and 0.35 respectively.  

\subsection{Qualitative Comparison} \label{sec:qualitative}
Figure \ref{fig:Examples} depicts the individuals containing the minimum, median and maximum novelty from each experiments' final populations (and seed population for comparison). By looking at these examples we can get a qualitative idea of how novelty and complexity is evolving over time. The seed population and static AE share similar high-level patterns which is understandable given they use the same autoencoder and originate from the same latent space.  The effect of the lack of training for the random AE experiment is clearly reflected in results which are far noisier than the other experiments.  The LS-AE, FH-AE and NA-AE experiments show slight differences in the overall structures generated compared to the seed set, though there is no significant jump in structural complexity.  This indicates that whilst novelty search is promoting diversity in the latent space, it does not guarantee diversity in the phenotype space and does not explicitly evolve towards desired qualities as in quality-diversity algorithms such as MAP-Elites \cite{mouret2015illuminating}.

\section{Discussion}\label{sec:discussion}
In this paper we apply novelty search in the latent space (with basic constraints) for the challenge of creative building generation in Minecraft. We tested three different approaches to the transformation of the autoencoder which differ in which evolved content is used for training. Results showed that autoencoders that used more past data (either as novelty archive or as final individuals of many exploration phases) were the most robust and best performers in all reconstruction error tests. The autoencoder trained on the latest set of results, on the other hand, seemed to overfit to the data and performed poorly on unseen data. The baseline which skipped the transformation phase altogether performed respectably across all the reconstruction tests, indicating the general patterns produced by the CPPNs remain concise even in later stages of evolution. The latest set autoencoder (LS-AE) and full history autoencoder (FH-AE) experiments had the least diverse final populations in the phenotype space, but produced the most different results compared to the seed populations. The novelty archive autoencoder (NA-AE) experiment maintained a population diversity on par with the baselines while also increasing diversity from the seed dataset. Finally, the NA-AE produced a latent distance function with the best correlation to our voxel-based diversity measure.

This line of research has focused on the creativity of the proposed generator; however, the lack of additional constraints to govern the feasibility of buildings is clear in the examples generated during our experiments (see Fig.~\ref{fig:Examples}). 
This issue could be mitigated through constraints enforcing basic architectural qualities (e.g., stability, symmetry, interior space requirements) and using a feasible-infeasible two-population \cite{kimbrough2008feasible} algorithm (FI-2POP) to improve the realism of the output. Using this approach, the repair functions could also be converted to feasibility constraints to allow the CPPNs to generate lattices end-to-end without the need for repair.
Another direction could be to seed the autoencoder or the initial population with pre-built structures (such as the medieval dataset used in this paper) to guide search towards house-like structures without imposing additional constraints and restricting creativity.  
Furthermore, the current approach does not create an appropriately designed interior space in the generated structures, simply carving out an enclosed volume and leaving it empty.  This can be fixed with a simple rule-based generator to design the interior space, or by incorporating a more complex interior generator into the pipeline in order to evolve the interior design in a creative fashion. Although the repair functions ensured that the buildings adhered to certain qualities, the current approach does not emphasize quality-diversity during evolution. Extending the current implementation with an algorithm such as MAP-Elites \cite{mouret2015illuminating} could help ensure diversity is preserved in desired qualities (e.g., stability, size) besides just novelty. Beyond these extensions, it is important to test the generator on higher resolution lattices, as the low resolution used in this paper is potentially limiting the visual impact of the output.

Latent vector evolution has most notably been applied to PCG for visuals \cite{hagg2021expressivity,nguyen2016understanding,zammit2022seeding}, level content \cite{Delenox}, robot behaviors \cite{cully2019autonomous}, as well as data efficient search space illumination \cite{gaier2018dataefficient}. To our knowledge, it has not been readily applied to constrained 3D structures, and so this paper opens up potential for further work in this domain. Our approach can also be applied to other facets within Minecraft and in new domains entirely. Broadly, it can be used in a mixed-initiative setting to promote creativity for level designers creating new content \cite{yannakakis2014mixed} by seeding the generator with human-designed content and generating novel suggestions. Within Minecraft, creative settlement generation (layout, adaptability, narrative) remains an open challenge presented by the GDMC report \cite{GDMC_Year1} which would benefit from open-ended evolution. Another interesting application would be generating agents with diverse/interesting behaviors which are evolved according to their surroundings. Our approach could also be adapted to other facets of computational creativity \cite{CGC}, such as the generation of audio files or game rules according to a set of constraints. 

\section{Conclusion}\label{sec:conclusions}
In this paper we proposed an autonomous building generator for Minecraft targeting open-ended complexity and creativity.  To achieve this the generator explores the problem space through CPPN-NEAT, evolving individuals according to novelty in the latent space, determined by a 3D autoencoder. The transformation phase retrains the autoencoder using high-performing (novel) examples from previous exploration phases to open up new areas of the search space and create more complex features. Whilst this work used very simple constraints for building generation, it provides a relatively unrestricted environment to test the creative capabilities of the generator. We tested various approaches for the transformation phase in order to identify its impact on population diversity and complexity over time, as well as the reconstruction accuracy of the autoencoders. Our results indicate that, by retraining the autoencoder, the generator is able to more effectively generate novel 3D structures compared to a static approach. The use of an autoencoder and CPPNs also ensures our approach is more scalable to higher resolution outputs, as the complexity of novelty search remains tied to the latent vector size. By addressing the limitations of our implementation, discussed above, we believe our approach can be an effective tool for generating diverse and creative content at a higher resolution, and with an open-ended complexity that is not currently achievable by the state of the art.

\section*{Acknowledgements}
This project has received funding from the European Union’s Horizon 2020 programme under grant agreement No 951911.

\bibliographystyle{IEEEtran}
\bibliography{delenox}
\end{document}